\documentclass{article}

\usepackage{microtype}
\usepackage{graphicx}
\usepackage{subfigure}
\usepackage{booktabs} 

\usepackage[backref]{hyperref}

\usepackage[accepted]{icml2023}

\usepackage{amsmath}
\usepackage{amssymb}
\usepackage{mathtools}
\usepackage{amsthm}
\usepackage{dsfont}
\usepackage{apxproof} 

\usepackage[capitalize,noabbrev]{cleveref}

\theoremstyle{plain}
\newtheorem{theorem}{Theorem}[section]
\newtheorem{proposition}[theorem]{Proposition}
\newtheorem{lemma}[theorem]{Lemma}
\newtheorem{corollary}[theorem]{Corollary}
\theoremstyle{definition}

\theoremstyle{remark}

\DeclareMathOperator*{\argmin}{arg\,min}
\usepackage{tabularx}
\usepackage[textsize=tiny]{todonotes}

\usepackage{xcolor}

\icmltitlerunning{Emergent Communication for Social Learning in MARL}

\begin{document}

\twocolumn[
\icmltitle{On the Role of Emergent Communication for Social Learning \\in Multi-Agent Reinforcement Learning
}



\icmlsetsymbol{equal}{*}

\begin{icmlauthorlist}
\icmlauthor{Seth Karten}{equal,yyy}
\icmlauthor{Siva Kailas}{yyy}
\icmlauthor{Huao Li}{pitt}
\icmlauthor{Katia Sycara}{yyy}
\end{icmlauthorlist}

\icmlaffiliation{yyy}{The Robotics Institute, Carnegie Mellon University, Pittsburgh, PA, USA}
\icmlaffiliation{pitt}{School of Information Science, University of Pittsburgh, Pittsburgh, PA, USA}

\icmlcorrespondingauthor{Seth Karten}{skarten@cs.cmu.edu}

\icmlkeywords{Multi-Agent Reinforcement Learning, Emergent Communication, Social Learning, Concept Whitening, Information Theory, Sparse Communication}

\vskip 0.3in
]



\printAffiliationsAndNotice{}  

\begin{abstract}

Explicit communication among humans is key to coordinating and learning. Social learning, which uses cues from experts, can greatly benefit from the usage of explicit communication to align heterogeneous policies, reduce sample complexity, and solve partially observable tasks. Emergent communication, a type of explicit communication, studies the creation of an artificial language to encode a high task-utility message directly from data. However, in most cases, emergent communication sends insufficiently compressed messages with little or null information, which also may not be understandable to a third-party listener. This paper proposes an unsupervised method based on the information bottleneck to capture both referential complexity and task-specific utility to adequately explore sparse social communication scenarios in multi-agent reinforcement learning (MARL). We show that our model is able to i) develop a natural-language-inspired lexicon of messages that is independently composed of a set of emergent concepts, which span the observations and intents with minimal bits, ii) develop communication to align the action policies of heterogeneous agents with dissimilar feature models, and iii) learn a communication policy from watching an expert's action policy, which we term `social shadowing'. 
\end{abstract}

\section{INTRODUCTION}

Social learning~\cite{jaques2019social,ndousse2021emergent} agents analyze cues from direct observation of other agents (novice or expert) in the same environment to learn an action policy from others.
However, observing expert actions may not be sufficient to coordinate with other agents.
Rather, by learning to communicate, agents can better model the intent of other agents, leading to better coordination.
In humans, explicit communication for coordination assumes a common communication substrate to convey abstract concepts and beliefs directly~\cite{mirsky2020penny}, which may not be available for new partners. 
To align complex beliefs, heterogeneous agents must learn a message policy that translates from one theory of mind~\cite{li2022theory} to another to synchronize coordination.
Especially when there is complex information to process and share, new agent partners need to learn to communicate to work with other agents.


Emergent communication studies the creation of artificial language. Often phrased as a Lewis game, speakers and listeners learn a set of tokens to communicate complex observations~\cite{lewis1969convention}. However, in multi-agent reinforcement learning (MARL), agents suffer from partial observability and non-stationarity (due to unaligned value functions)~\cite{papoudakis2019dealing}, which aims to be solved with decentralized learning through communication. In the MARL setup, agents, as speakers and listeners, learn a set of tokens to communicate observations, intentions, coordination, or other experiences which help facilitate solving tasks~\cite{karten2022sparse,karten2022inter}. Agents learn to communicate effectively through a backpropagation signal from their task performance~\cite{foerster2016learning, lowe2017multi, lazaridou2016multi, commnet, ic3net}.
This has been found useful for applications in human-agent teaming~\cite{karten2022inter,marathe2018bidirectional,lake2019human,lazaridou2020emergent}, multi-robot navigation~\cite{benSparseDiscrete}, and coordination in complex games such as StarCraft II~\cite{samvelyan2019starcraft}.
Communication quality has been shown to have a strong relationship with task performance~\cite{marlow2018does}, leading to a multitude of work attempting to increase the representational capacity by decreasing the convergence rates~\cite{EcclesBiases,MA_autoencoder,karten2022sparse,wang2020learning,tucker2022towards}. Yet these methods still create degenerate communication protocols~\cite{karten2022inter,karten2022sparse,benSparseDiscrete}, which are uninterpretable due to joined concepts or null (lack of) information, which causes performance degradation.

In this work, we investigate the challenges of learning a messaging lexicon to prepare emergent communication for social learning (EC4SL) scenarios.
We study the following hypotheses:
\textbf{H1)} EC4SL will learn faster through structured concepts in messages leading to higher-quality solutions,
\textbf{H2)} EC4SL aligns the policies of expert heterogeneous agents,
and \textbf{H3)} EC4SL enables social shadowing, where an agent learns a communication policy  while only observing an expert agent's action policy. By learning a communication policy, the agent is encouraged to develop a more structured understanding of intent, leading to better coordination.
The setting is very realistic among humans and many computer vision and RL frameworks may develop rich feature spaces for a specific solo task, but have not yet interacted with other agents, which may lead to failure without alignment.

We enable a compositional emergent communication paradigm, which exhibits clustering and informativeness properties.
We show theoretically and through empirical results that compositional language enables independence properties among tokens with respect to referential information. Additionally, when combined with contrastive learning, our method outperforms competing methods that only ground communication on referential information. We show that contrastive learning is an optimal critic for communication, reducing sample complexity for the unsupervised emergent communication objective.
In addition to the more human-like format, compositional communication is able to create variable-length messages, meaning that we are not limited to sending insufficiently compressed messages with little information, increasing the quality of each communication.

In order to test our hypotheses, we show the utility of our method in multi-agent settings with a focus on teams of agents, high-dimensional pixel data, and expansions to heterogeneous teams of agents of varying skill levels.
Social learning requires agents to explore to observe and learn from expert cues. 
We interpolate between this form of social learning and imitation learning, which learns action policies directly from examples. 
We introduce a 'social shadowing' learning approach where we use first-person observations, rather than third-person observations, to encourage the novice to learn latently or conceptually how to communicate and develop an understanding of intent for better coordination. The social shadowing episodes are alternated with traditional MARL during training.
Contrastive learning, which works best with positive examples, is apt for social shadowing.
Originally derived to enable lower complexity emergent lexicons, we find that the contrastive learning objective is apt for agents to develop internal models and relationships of the task through social shadowing.

The idea is to enable a shared emergent communication substrate (with minimal bandwidth) to enable future coordination with novel partners. Our contributions are deriving an optimal critic for a communication policy and showing that the information bottleneck helps extend communication to social learning scenarios. In real-world tasks such as autonomous driving or robotics, humans do not necessarily learn from scratch. Rather they explore with conceptually guided information from expert mentors. In particular, having structured emergent messages reduces sample complexity, and contrastive learning can help novice agents learn from experts.  Emergent communication can also align heterogeneous agents, a social task that has not been previously studied.

\section{RELATED WORK}


\subsection{Multi-Agent Signaling}
Implicit communication conveys information to other agents that is not intentionally communicated~\cite{grupen2022multi}. 
Implicit signaling conveys information to other agents based on one's observable physical position~\cite{grupen2022multi}. Implicit signaling may be a form of implicit communication such as through social cues~\cite{jaques2019social,ndousse2021emergent} or explicit communication such as encoded into the MDP through ``cheap talk"~\cite{sokota2022communicating}.
Unlike implicit signaling, explicit signaling is a form of positive signaling~\cite{li2021learning} that seeks to directly influence the behavior of other agents in the hopes that the new information will lead to active listening.
Multi-agent emergent communication is a type of explicit signaling which deliberately shares information.
Symbolic communication, a subset of explicit communication, seeks to send a subset of pre-defined messages. However, these symbols must be defined by an expert and do not scale to particularly complex observations and a large number of agents.
Emergent communication aims to directly influence other agents with a learned subset of information, which allows for scalability and interpretability by new agents.


\subsection{Emergent Communication}
Several methodologies currently exist to increase the informativeness of emergent communication.
With discrete and clustered continuous communication, the number of observed distinct communication tokens is far below the number permissible~\cite{discreteComm}.
As an attempt to increase the emergent ``vocabulary'' and decrease the data required to converge to an informative communication ``language'', work has added a bias loss to emit distinct tokens in different situations~\cite{EcclesBiases}.
More recent work has found that the sample efficiency can be further improved by grounding communication in observation space with a supervised reconstruction loss~\cite{MA_autoencoder}.
Information-maximizing autoencoders aim to maximize the state reconstruction accuracy for each agent.
However, grounding communication in observations has been found to easily satisfy these input-based objectives while still requiring a myriad more samples to explore to find a task-specific communication space~\cite{karten2022sparse}. Thus, it is necessary to use task-specific information to communicate informatively. This will enable learned compression for task completion rather than pure compression for input recovery.
Other work aims to use the information bottleneck~\cite{tishby2015deep} to decrease the entropy of messages~\cite{wang2020learning}. 
In our work, we use contrastive learning to increase representation similarity with future goals, which we show optimally optimizes the Q-function for messages. 

\subsection{Natural Language Inspiration}
The properties of the tokens in emergent communication directly affect their informative ability.
As a baseline, continuous communication tokens can represent maximum information but lack human-interpretable properties.
Discrete 1-hot (binary vector) tokens allow for a finite vocabulary, but each token contains the same magnitude of information, with equal orthogonal distance to each other token.
Similar to word embeddings in natural language, discrete prototypes are an effort to cluster similar information together from continuous vectors~\cite{discreteComm}. Building on the continuous word embedding properties, VQ-VIB~\cite{tucker2022towards}, an information-theoretic observation grounding based on VQ-VAE properties~\cite{van2017neural}, uses variational properties to provide word embedding properties for continuous emergent tokens. Like discrete prototypes, they exhibit a clustering property based on similar information but are more informative.
However, each of these message types determines a single token for communication. 
Tokens are stringed together to create emergent ``sentences''.




\section{Preliminaries}

We formulate our setup as a decentralized, partially observable Markov Decision Process with communication (Dec-POMDP-Comm).
Formally, our problem is defined by the tuple, $\langle\mathcal{S},\mathcal{A},\mathcal{M},\mathcal{T},\mathcal{R},\mathcal{O},\Omega,\gamma \rangle$. We define $\mathcal{S}$ as the set of states, $\mathcal{A}^i \, , \, i\in[1,N]$ as the set of actions, which includes task-specific actions, and $\mathcal{M}^i$ as the set of communications for $N$ agents.  $\mathcal{T}$ is the transition between states due to the multi-agent joint action space $\mathcal{T}: \mathcal{S} \times \mathcal{A}^1,...,\mathcal{A}^N \to \mathcal{S}$. $\Omega$ defines the set of observations in our partially observable setting. Partial observability requires communication to complete the tasks successfully. $\mathcal{O}^i: \mathcal{M}^1,...,\mathcal{M}^N \times \hat{\mathcal{S}} \to \Omega$ maps the communications and local state, $\hat{\mathcal{S}}$, to a distribution of observations for each agent. $\mathcal{R}$ defines the reward function and $\gamma$ defines the discount factor.

\subsection{Architecture}
The policy network is defined by three stages: Observation Encoding, Communication, and Action Decoding. The best observation encoding and action decoding architecture is task-dependent, i.e., using multi-layer perceptrons (MLPs), CNNs~\cite{lecun1995convolutional}, GRUs~\cite{chung2014empirical}, or transformer~\cite{vaswani2017attention} layers are best suited to different inputs. 
The encoder transforms observation and any sequence or memory information into an encoding $H$. The on-policy reinforcement learning training uses REINFORCE~\cite{williams1992simple} or a decentralized version of MAPPO~\cite{yu2021surprising} as specified by our experiments.

Our work focuses on the communication stage, which can be divided into three substages: message encoding, message passing (often considered sparse communication), and message decoding. We use the message passing from~\cite{karten2022sparse}. For message decoding, we build on a multi-headed attention framework, which allows an agent to learn which messages are most important~\cite{graphMA}. Our compositional communication framework defines the message encoding, as described in section~\ref{sec:composition}.

\subsection{Objective}
Mutual information, denoted as $I(X;Y)$, looks to measure the relationship between random variables,
\begin{equation*}
    \begin{aligned}
        I(X;Y) = \mathds{E}_{p(x,y)} \left[ \log\frac{p(x|y)}{p(x)} \right] = \mathds{E}_{p(x,y)} \left[ \log\frac{p(y|x)}{p(y)} \right]
    \end{aligned}
\end{equation*}
which is often measured through Kullback-Leibler divergence~\cite{kullback1997information}, $I(X;Y) = D_{KL} (p(x,y) || p(x) \otimes p(y))$. 
The message encoding substage can be defined as an information bottleneck problem, which defines a trade-off between the complexity of information (compression, $I(X,\hat{X})$) and the preserved relevant information (utility, $I(\hat{X},Y)$).
The deep variational information bottleneck defines a trade-off between preserving useful information and compression~\cite{alemi2017deep,tishby2015deep}.
We assume that our observation and memory/sequence encoder provides an optimal representation $H^i$ suitable for sharing relevant observation and intent/coordination information. We hope to recover a representation $Y^i$, which contains the sufficient desired outputs.

In our scenario, the information bottleneck is a trade-off between the complexity of information $I(H^i; M^i)$ (representing the encoded information exactly) and representing the relevant information $I(M^{j \neq i}; Y^i) $, which is signaled from our contrastive objective. In our setup, the relevant information flows from other agents through communication, signaling a combination of the information bottleneck and a Lewis game. We additionally promote complexity through our compositional independence objective, $I(M_1^i;\hdots ; M_L^i | H^i) $. This is formulated by the following Lagrangian,
\begin{align*}
    \mathcal{L}(\ p(m^i | h^i )\ ) =\ &\beta_u \hat{I} (M^{j \neq i}; Y^i)\ - \beta_c \hat{I}(H^i; M^i) \\&- \beta_I \hat{I} (M_1^i;\hdots ; M_L^i | H^i) 
\end{align*}
where the bounds on mutual information $\hat{I}$ are defined in equations~\ref{eq:inde_info},~\ref{eq:input}, and~\ref{eq:contrastive}. 
Overall, our objective is,
\begin{align*}
    J(\theta) = \max\limits_{\pi} \mathds{E} \left[ \sum_{t \in T} \sum_{i \in N} \gamma_t \mathcal{R}(s_t,a_t) + \mathcal{L}(\ p(m_t | h_t )\ )  \right] \\
    \text{s.t.} (a_t, m_t, h_t) \sim \pi^i, s_t \sim \mathcal{T}(s_{t-1})
\end{align*}

\section{Complexity through Compositional Communication}\label{sec:composition}

We aim to satisfy the complexity objective, $I(H^i, M^i)$, through compositional communication.
In order to induce complexity in our communication, we want the messages to be as non-random as possible. That is, informative with respect to the input hidden state $h$. In addition, we want each token within the message to share as little information as possible with the preceding tokens. Thus, each additional token adds \textit{only informative} content. Each token has a fixed length in bits $W$. The total sequence is limited by a fixed limit, $\sum_l^L W_l \leq S$, of $S$ bits and a total of $L$ tokens. 

We use a variational message generation setup, which maps the encoded hidden state $h$ to a message $m$; that is, we are modeling the posterior, $\pi_m^i (m_l|h)$. We limit the vocabulary size to $K$ tokens, $e_j \in \mathds{R}^D, j \in [1,K] \subset \mathds{N}$, where each token has dimensionality $D$  and $l \in [1,L] \subset \mathds{N}$. Each token $m_l$ is sampled from a categorical posterior distribution,
\begin{equation*}
    \pi_m^i (m_l = e_k | h) = 
    \begin{cases}
        1 & \text{for } k = \argmin\limits_{j} || m_l - e_j ||_2 \\
        0 & \text{otherwise}
    \end{cases}
\end{equation*}
such that the message $m_l$ is mapped to the nearest neighbor $e_j$. A set of these tokens makes a message $m$.
To satisfy the complexity objective, we want to use $m^i$ to well-represent $h^i$ and consist of independently informative $m^i_l$.

\subsection{Independent Information}
We derive an upper bound for the interaction information between all tokens.
\begin{proposition}
For the interaction information between all tokens, the following upper bound holds: $I(m_1; \hdots; m_L | h) \leq \mathds{E}_{h \sim p(h)} \left[ D_{KL} \left(q(\hat{m}|h) || \pi^i_m(m_1|h) \otimes \cdots \otimes \pi^i_m(m_L|h)\right) \right]$.
\end{proposition}
The proof is in Appendix~\ref{appx:proofs}.


\begin{figure*}[!t]
    \centering
    \includegraphics[width=.75\textwidth]{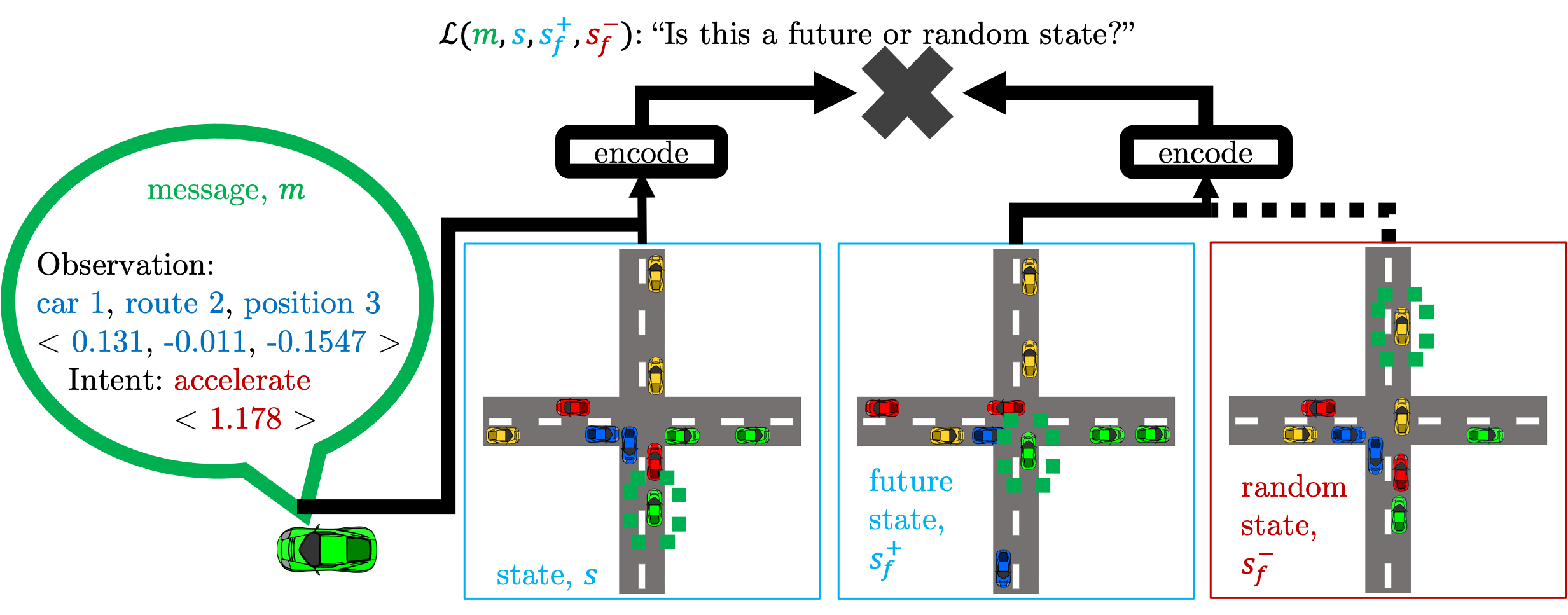}
    \caption{By using contrastive learning, our method seeks similar representations between the state-message pair and future states while creating dissimilar representations with random states. Thus satisfying the utility objective of the information bottleneck. The depicted agents are blind and cannot see other cars.}
    \label{fig:fig2}
\end{figure*}


Since we want the mutual information to be minimized in our objective, we minimize,
\begin{equation}\label{eq:inde_info}
\begin{aligned}
    &\hat{I} (m_1;\hdots ; m_L | h) =\\
    &\mathds{E}_{h \sim p(h)} \left[ D_{KL} \left(q(\hat{m}|h) || \pi^i_m(m_1|h) \otimes \cdots \otimes \pi^i_m(m_L|h)\right) \right]    
\end{aligned}
\end{equation}

\subsection{Input-Oriented Information}

In order to induce complexity in the compositional messages, we additionally want to minimize the mutual information $I(H; M)$ between the composed message $\hat{m}$ and the encoded information $h$.
We derive an upper bound on the mutual information that we use as a Lagrangian term to minimize.
\begin{proposition}
For the mutual information between the composed message and encoded information, the following upper bound holds: $I(H; M) \leq \sum_l^L \mathds{E}_{h \sim p(h)} \left[ D_{KL} \left( q(m_l|h) || z(m_l)) \right) \right]$.
\end{proposition}
The proof is in Appendix~\ref{appx:proofs}. Thus, we have our Lagrangian term,
\begin{equation}\label{eq:input}
\begin{aligned}
    \hat{I}(H^i, M^i) = \sum_l^L \mathds{E}_{h \sim p(h)} \left[ D_{KL} \left( q(m_l|h) || z(m_l)) \right) \right]
\end{aligned}
\end{equation}

Conditioning on the input or observation data is a decentralized training objective.

\subsection{Sequence Length}
Compositional communication necessitates an adaptive limit on the total length of the sequence. 
\begin{corollary}\label{cor:redundant}
    Repeat tokens, $w$, are redundant and can be removed.
\end{corollary}
Suppose one predicts two arbitrary tokens, $w_k$ and $w_l$. Given equation~\ref{eq:inde_info}, it follows that there is low or near-zero mutual information between $w_k$ and $w_l$.

A trivial issue is that the message generator will predict every available token as to follow the unique token objective. Since the tokens are imbued with input-oriented information (equation~\ref{eq:input}), the predicted tokens will be based on relevant referential details. Thus, it follows that tokens containing irrelevant information will not be chosen.

A nice optimization objective that follows from corollary~\ref{cor:redundant} is that one can use self-supervised learning with an end-of-sequence (EOS) token to limit the variable total length of compositional message sequences.

\begin{equation}\label{eq:seq_len}
    H(m_{\texttt{EOS}}, m_l) = - \pi(m_{\texttt{EOS}}) \log(\pi(m_l))
\end{equation}

\subsection{Message Generation Architecture}
Now, we can define the pipeline for message generation. The idea is to create an architecture that can generate features to enable independent message tokens.
We expand each compressed token into the space of the hidden state $h$ (1-layer linear expansion) since each token has a natural embedding in $\mathbf{R}^{|h|}$. 
Then, we perform attention using a \texttt{softmin} to help minimize similarity with previous tokens and sample the new token from a variational distribution.
See algorithm~\ref{alg:cap} for complete details.
During execution, we can generate messages directly due to equation~\ref{eq:inde_info}, resolving any computation time lost from sequential compositional message generation.

\begin{algorithm}[!t]
\caption{\texttt{Compositional Message Gen.}$(h_t)$}\label{alg:cap}
\begin{algorithmic}[1]
\STATE $T \gets \texttt{num\_tokens}$
\STATE $m = \textbf{0}$ \COMMENT{$T \times d_m$, $d_m \gets \texttt{token\_size}$}
\STATE $Q \gets \texttt{Q\_MLP}(h_t)$
\STATE $V \gets \texttt{V\_MLP}(h_t)$
\FOR{$i \gets 1 \text{ to } T$}
    \STATE $K \gets \texttt{K\_MLP}(m)$
    \STATE $\hat{h} = \texttt{softmin}(\frac{Q^\intercal \texttt{mean}(K,1)}{\sqrt{d_k}})^\intercal V$
    
    \STATE $m_i \sim \mathcal{N}(\hat{h}; \mu, \sigma)$
\ENDFOR
\STATE \textbf{return} $m$
\end{algorithmic}
\end{algorithm}  
\section{Utility through Contrastive Learning}

First, note that our Markov Network is as follows: $H^j \rightarrow M^j \rightarrow Y^i \leftarrow H^i$. Continue to denote $i$ as the agent identification and $j$ as the agent ID such that $j \neq i$.
We aim to satisfy the utility objective of the information bottleneck, $I(M^j; Y^i)$, through contrastive learning as shown in figure~\ref{fig:fig2}. 
\begin{proposition}
Utility mutual information is lower bounded by the contrastive NCE-binary objective, $I(M,Y) \geq \log \sigma (f(s,m,s_f^+)) +\log \sigma (1-f(s,m,s_f^-))$.
\end{proposition}

The proof is in Appendix~\ref{appx:proofs}.


This result shows a need for gradient information to flow backward across agents along communication edge connections.


\section{Experiments and Results}
We condition on inputs, especially rich information (such as pixel data), and task-specific information.
When evaluating an artificial language in MARL, we are interested in referential tasks, in which communication is \textit{required} to complete the task.
With regard to intent-grounded communication, we study ordinal tasks, which require coordination information between agents to complete successfully.
Thus, we consider tasks with a team of agents to foster messaging that communicates coordination information that also includes their observations.
To test \textbf{H1}, structuring emergent messages enables lower complexity, we test our methodology and analyze the input-oriented information and utility capabilities.
Next, we analyze the ability of heterogeneous agents to understand differing communication policies (\textbf{H2})).
Finally, we consider the effect of social shadowing (\textbf{H3}), in which agents solely learn a communication policy from an expert agent's action policy.
We additionally analyze the role of offline reinforcement learning for emergent communication in combination with online reinforcement learning to further learn emergent communication alongside an action policy. 
We evaluate each scenario over 10 seeds.

\begin{figure}[!t]
    \centering
    \includegraphics[width=0.5\columnwidth]{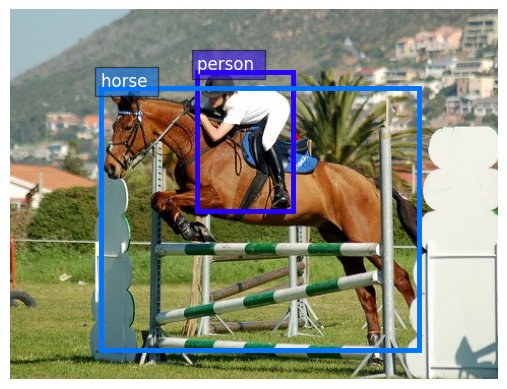}
    \caption{An example of two possible classes, person and horse, from a single observation in the Pascal VOC game.}
    \label{fig:pascal_ex}
\end{figure}

\subsection{Environments}

\paragraph{Blind Traffic Junction} We consider a benchmark that requires both referential and ordinal capabilities within a team of agents. The blind traffic junction environment~\cite{ic3net} requires multiple agents to navigate a junction without any observation of other agents. Rather, they only observe their own state location. Ten agents must coordinate to traverse through the lanes without colliding into agents within their lane or in the junction. Our training uses REINFORCE~\cite{williams1992simple}.

\paragraph{Pascal VOC Game} We further evaluate the complexity of compositional communication with a Pascal VOC~\cite{everingham2010pascal}. This is a two-agent referential game similar to the Cifar game~\cite{MA_autoencoder} but requires the prediction of multiple classes. During each episode, each agent observes a random image from the Pascal VOC dataset containing exactly two unique labels. Each agent must encode information given only the raw pixels from the original image such that the other agent can recognize the two class labels in the original image. An agent receives a reward of 0.25 per correctly chosen class label and will receive a total reward of 1 if both agents guess all labels correctly. See figure~\ref{fig:pascal_ex}. Our training uses heterogeneous agents trained with PPO (modified from MAPPO~\cite{yu2021surprising} repository).
For simplicity of setup, we consider images with exactly two unique labels from a closed subset of size five labels of the original set of labels from the Pascal VOC data. Furthermore, these images must be of size $375 \times 500$ pixels. Thus, the resultant dataset comprised 534 unique images from the Pascal VOC dataset. 


\subsection{Baselines}
To evaluate our methodology, we compare our method 
to the following baselines:
(1) \texttt{no-comm}, where agents do not communicate;
(2) \texttt{rl-comm}, which uses a baseline communication method learned solely through policy loss~\cite{ic3net};
(3) \texttt{ae-comm}, which uses an autoencoder to ground communication in input observations~\cite{MA_autoencoder};
(4) \texttt{VQ-VIB}, which uses a variational autoencoder to ground discrete communication in input observations and a mutual information objective to ensure low entropy communication~\cite{tucker2022towards}.




\begin{table}[t!]
\centering
\caption{Beta ablation: Messages are naturally sparse in bits due to the complexity loss. Redundancy measures the capacity for a bijection between the size of the set of unique tokens and the enumerated observations and intents. Min redundancy is 1.0 (a bijection). Lower is better.}
\begin{tabularx}{\columnwidth}{|m{.22\columnwidth}|m{.18\columnwidth} m{.2\columnwidth} m{.19\columnwidth} |}
 \specialrule{.2em}{.1em}{.1em} 
   $\beta$ & Success & Message Size in Bits & Redundancy \\\hline
   0.1 & 1.0 & 64 & 1.0 \\
   0.01 & .996 & 69.52 & 1.06 \\
   0.001 & .986 & 121.66 & 2.06 \\
   0 & .976 & 147.96 & 2.31 \\
   non-compositional & .822 & 512 & 587 \\
\specialrule{.1em}{.05em}{.05em}
\end{tabularx}
\label{table:beta}
\end{table}

\subsection{Input-Oriented Information Results}
We provide an ablation of the loss parameter $\beta$ in table~\ref{table:beta} in the blind traffic junction scenario. When $\beta = 0$, we use our compositional message paradigm without our derived loss terms.
We find that higher complexity and independence losses increase sample complexity. When $\beta=1$, the model was unable to converge. However, when there is no regularization loss, the model performs worse (with no guarantees about referential representation). We attribute this to the fact that our independence criteria learns a stronger causal relationship. There are fewer spurious features that may cause an agent to take an incorrect action.


In order to understand the effect of the independent concept representation, we analyze the emergent language's capacity for redundancy. A message token $m_l$ is redundant if there exists another token $m_k$ that represents the same information. 
With our methodology, the emergent `language' converges to the exact number of observations and intents required to solve the task. 
With a soft discrete threshold, the independent information loss naturally converges to a discrete number of tokens in the vocabulary.
Our $\beta$ ablation in table~\ref{table:beta} yields a bijection between each token in the vocabulary and the possible emergent concepts, i.e., the enumerated observations and intents. Thus for $\beta = 0.1$, there is no redundancy.

\paragraph{Sparse Communication} In corollary~\ref{cor:redundant}, we assume that there is no mutual information between tokens. In practice, the loss may only be near-zero. Our empirical results yield independence loss around $1e-4$. In table~\ref{table:beta}, the size of the messages is automatically compressed to the smallest size to represent the information. Despite a trivially small amount of mutual information between tokens, our compositional method is able to reduce the message size in bits by 2.3x using our derived regularization, for a total of an 8x reduction in message size over non-compositional methods such as $\texttt{ae-comm}$. Since the base unit for the token is a 32-bit float, we note that each token in the message may be further compressed. We observe that each token uses three significant digits, which may further compress tokens to 10 bits each for a total message length of 20 bits. 

\begin{figure}[!t]
    \centering
    \includegraphics[width=0.49\columnwidth]{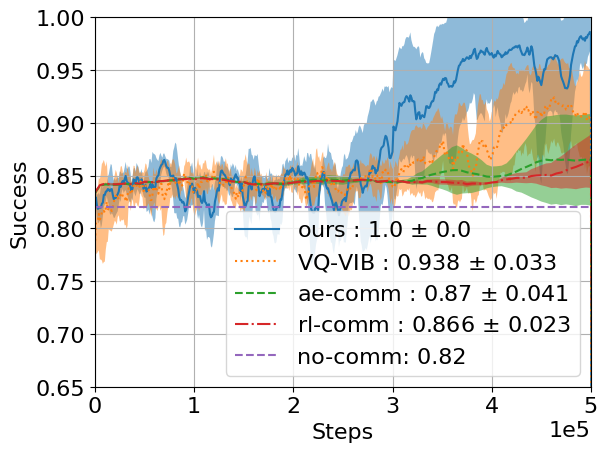}
    \includegraphics[width=0.49\columnwidth]{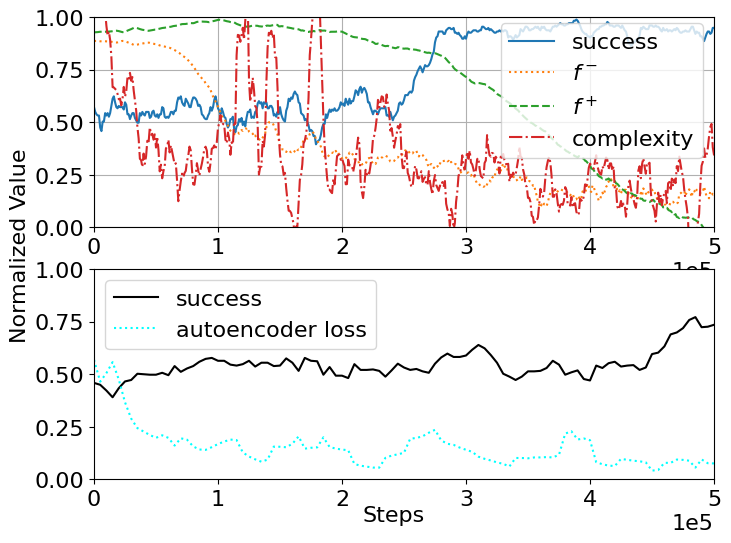}
    \caption{\textbf{Blind Traffic Junction} Left: Our method uses compositional complexity and contrastive utility to outperform other baselines in terms of performance and sample complexity. The legend provides the mean $\pm$ variance of the best performance. Right: Top: success, contrastive, and complexity losses for our method. Right, Bottom: success, autoencoder loss for \texttt{ae-comm} with supervised pretraining. }
    \label{fig:contrastive}
    \label{fig:loss_anal}
\end{figure}


\subsection{Communication Utility Results}
Due to coordination in MARL, grounding communication in referential features is not enough. Finding the communication utility requires grounding messages in ordinal information.
Overall, figure~\ref{fig:contrastive} shows that our compositional, contrastive method outperforms all methods focused on solely input-oriented communication grounding. In the blind traffic junction, our method yields a higher average task success rate and is able to achieve it with a lower sample complexity.
Training with the contrastive update tends to spike to high success but not converge, often many episodes before convergence, which leaves area for training improvement. That is, the contrastive update begins to find aligned latent spaces early in training, but it cannot adapt the methodology quickly enough to converge. The exploratory randomness of most of the early online data prevents exploitation of the high utility $f^+$ examples. This leaves further room for improvement for an adaptive contrastive loss term.

\paragraph{Regularization loss convergence} After convergence to high task performance, the autoencoder loss increases in order to represent the coordination information. This follows directly from the information bottleneck, where there exists a tradeoff between utility and complexity. However, communication, especially referential communication, should have an overlap between utility and complexity. Thus, we should seek to make the complexity loss more convex. Our compositional communication complexity loss does not converge before task performance convergence. While the complexity loss tends to spike in the exploratory phase, the normalized value is very small. Interestingly, the method eventually converges as the complexity loss converges below a normalized 0.3. Additionally, the contrastive loss tends to decrease monotonically and converges after the task performance converges, showing a very smooth decrease. The contrastive $f^-$ loss decreases during training, which may account for success spikes prior to convergence. The method is able to converge after only a moderate decrease in the $f^+$ loss.
This implies empirical evidence that the contrastive loss is an optimal critic for messaging. See figure~\ref{fig:loss_anal}.

\begin{figure}[!t]
    \centering
    \includegraphics[width=.75\columnwidth]{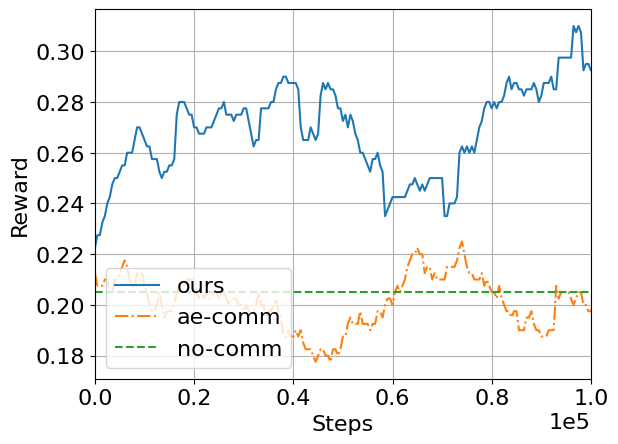}
    \caption{\textbf{Pascal VOC Game} Representing compositional concepts from raw pixel data in images to communicate multiple concepts within a single image. Our method significantly outperforms \texttt{ae-comm} and \texttt{no-comm} due to our framework being able to learn composable, independent concepts.}
    \label{fig:pascal}
\end{figure}

\subsection{Heterogeneous Alignment Through Communication}
In order to test the heterogeneous alignment ability of our methodology to learn higher-order concepts from high-dimensional data, we analyze the performance on the Pascal VOC game.
We compare our methodology against \texttt{ae-comm} to show that concepts should consist of independent information directly from task signal rather than compression to reconstruct inputs. That is, we show an empirical result on pixel data to verify the premise of the information bottleneck.
Our methodology significantly outperforms the observation-grounded \texttt{ae-comm} baseline, as demonstrated by figure~\ref{fig:pascal}. The \texttt{ae-comm} methodology, despite using autoencoders to learn observation-grounded communication, performs only slightly better than \texttt{no-comm}. On the other hand, our methodology is able to outperform both baselines significantly. It is important to note that based on figure~\ref{fig:pascal}, our methodology is able to guess more than two of the four labels correctly across the two agents involved, while the baseline methodologies struggle to guess exactly two of thew four labels consistently. This can be attributed to our framework being able to learn compositional concepts that are much more easily discriminated due to mutual independence.

\subsection{Social Shadowing}
\begin{figure}[!t]
    \centering
    \includegraphics[width=.75\columnwidth]{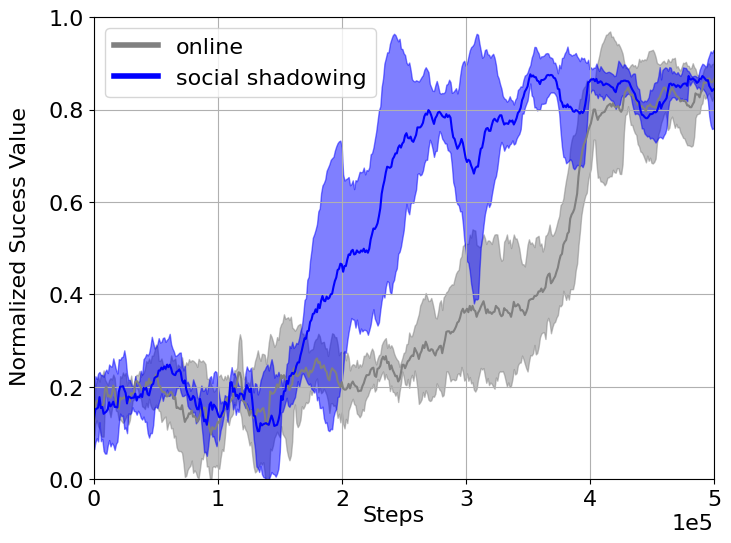}
    \caption{\textbf{Blind Traffic Junction} Social shadowing enables significantly lower sample complexity when compared to traditional online MARL.}
    \label{fig:tj_social_teach}
\end{figure}

Critics of emergent communication may point to the increased sample complexity due to the dual communication and action policy learning.
In the social shadowing scenario, heterogeneous agents can learn to generate a communication policy without learning the action policy of the watched expert agents.
To enable social shadowing, the agent will alternate between a batch of traditional MARL (no expert) and (1st-person) shadowing an expert agent performing the task in its trajectory. The agent only uses the contrastive objective to update its communication policy during shadowing.
In figure~\ref{fig:tj_social_teach}, the agent that performs social shadowing is able to learn the action policy with almost half the sample complexity required by the online reinforcement learning agent.
Our results show that the structured latent space of the emergent communication learns socially benevolent coordination. This tests our hypothesis that by learning communication to understand the actions of other agents, one can enable lower sample complexity coordination. Thus, it mitigates the issues of solely observing actions.

\section{Discussion}
By using our framework to better understand the intent of others, agents can learn to communicate to align policies and coordinate.
Any referential-based setup can be performed with a supervised loss, as indicated by the instant satisfaction of referential objectives. Even in the Pascal VOC game, which appears to be a purely referential objective, our results show that intelligent compression is not the only objective of referential communication. The emergent communication paradigm must enable an easy-to-discriminate space for the game. In multi-agent settings, the harder challenge is to enable coordination through communication. Using contrastive communication as an optimal critic aims to satisfy this, and has shown solid improvements.
Since contrastive learning benefits from good examples, this method is even more powerful when there is access to examples from expert agents. In this setting, the communication may be bootstrapped, since our optimal critic has examples with strong signals from the 'social shadowing' episodes. 

Additionally, we show that the minimization of our independence objective enables tokens that contain minimal overlapping information with other tokens. Preventing trivial communication paradigms enables higher performance. Each of these objectives is complementary, so they are not trivially minimized during training, which is a substantial advantage over comparative baselines. Unlike prior work, this enables the benefits of training with reinforcement learning in multi-agent settings. 

In addition to lower sample complexity, the mutual information regularization yields additional benefits, such as small messages, which enables the compression aspect of sparse communication. From a qualitative point of view, the independent information also yields discrete emergent concepts, which can be further made human-interpretable by a post-hoc analysis~\cite{yeh2021human}. This is a step towards white-box machine learning in multi-agent settings. The interpretability of this learned white-box method could be useful in human-agent teaming as indicated by prior work~\cite{karten2022inter}. The work here will enable further results in decision-making from high-dimensional data with emergent concepts.
The social scenarios described are a step towards enabling a zero-shot communication policy. This work will serve as future inspiration for using emergent communication to enable ad-hoc teaming with both agents and humans.



\bibliography{root}
\bibliographystyle{icml2023}

\newpage
\appendix
\onecolumn
\section{Appendix}

\subsection{Proofs}\label{appx:proofs}

\textbf{Proposition 4.1} \textit{For the interaction information between all tokens, the following upper bound holds: $I(m_1; \hdots; m_L | h) \leq \mathds{E}_{h \sim p(h)} \left[ D_{KL} \left(q(\hat{m}|h) || \pi^i_m(m_1|h) \otimes \cdots \otimes \pi^i_m(m_L|h)\right) \right]$.}
\begin{proof}
Starting with the independent information objective, we want to minimize the interaction information,
\begin{equation*}
    \begin{aligned}
        &I(m_1; \hdots; m_L | h) = \\
        &\int \hdots \int f_m(m_1, \hdots, m_L, h) dh\ d{m_1} \hdots d{m_L}
    \end{aligned}
\end{equation*}
which defines the conditional mutual information between each token and,
\begin{equation}    \label{eq:interaction}
    \begin{aligned}
        f_m(*) = p(h) p(m_1; \hdots; m_L | h) \log \frac{p(m_1; \hdots; m_L | h)}{\prod_l^L p(m_L | h)} 
    \end{aligned}
\end{equation}

Let $\pi_m^i (m_l | h)$ be a variational approximation of $p(m_l | h)$, which is defined by our message encoder network. 
Given that each token should provide unique information, we assume independence between $m_l$. Thus, it follows that our compositional message is a vector, $m = [m_1, \hdots, m_L]$, and is jointly Gaussian. Moreover, we can define $q(\hat{m} | h)$ as a variational approximation to $p(m | h) = p(m_1; \hdots, m_L | h)$. We can model $q$ with a network layer and define its loss as $||\hat{m} - m||_2$.
Thus, transforming equation \ref{eq:interaction} into variational form, we have,
\begin{equation*}
    \begin{aligned}
        g_m(m_1, \hdots, m_L, h) = 
        p(h) q(\hat{m} | h) \log \frac{q(\hat{m}|h)}{\prod_l^L \pi_m^i(m_l | h)}
    \end{aligned}
\end{equation*}

Since Kullback-Leibler divergence $D_{KL}$ is non-negative, 
$$D_{KL}\left(q(\hat{m}|h) || \pi^i_m(m_1|h) \otimes \cdots \otimes \pi^i_m(m_L|h)\right) \geq 0,$$ 
it follows that 
$$\int q(\hat{m}|h) \log q(\hat{m}|h) d\hat{m} \geq \int q(\hat{m}|h) \log \prod_l^L \pi^i_m(m_l|h) d\hat{m}$$
Thus, we can bound our interaction information,
\begin{equation*}
    \begin{aligned}
        &I(m_1;\hdots ; m_L | h) \leq \int \hdots \int g_m(*) dh d{m_1} \hdots d{m_L} \\
        &= \mathds{E}_{h \sim p(h)} \left[ D_{KL} \left(q(\hat{m}|h) || \pi^i_m(m_1|h) \otimes \cdots \otimes \pi^i_m(m_L|h)\right) \right]
    \end{aligned}
\end{equation*}
\end{proof}

\textbf{Proposition 4.2} \textit{For the mutual information between the composed message and encoded information, the following upper bound holds: $I(H; M) \leq \sum_l^L \mathds{E}_{h \sim p(h)} \left[ D_{KL} \left( q(m_l|h) || z(m_l)) \right) \right]$.}
\begin{proof}
By definition of mutual information between the composed messages $M$ and the encoded observations $H$, we have,
\begin{equation*}
    \begin{aligned}
        I(H; M) = \int \int p(h) p(\hat{m}|h) \log \frac{p(\hat{m}|h)}{p(\hat{m})} d\hat{m}\ dh
    \end{aligned}
\end{equation*}
Substituting $q(\hat{m}|h)$ for $p(\hat{m}|h)$, the same KL Divergence identity, and defining a Gaussian approximation $z(\hat{m})$ of the marginal distribution $p(\hat{m})$, it follows that,
\begin{equation*}
    \begin{aligned}
        I(H; M) \leq \int \int p(h) q(\hat{m}|h) \log \frac{q(\hat{m}|h)}{z(\hat{m})} d\hat{m}\ dh
    \end{aligned}
\end{equation*}
In expectation of equation~\ref{eq:inde_info}, we have, $$q(\hat{m}|h) = q(\hat{m}|h) = \prod_l^L \pi^i_m (m_l|h).$$
This implies that, for $\hat{m}=[m_1,\hdots,m_L]$, there is probabilistic independence between $m_j, m_k, j\neq k$.
Thus, expanding, it follows that,
\begin{equation*}
    \begin{aligned}
        I(H; M) &\leq \sum_l^L \int \int p(h) q(m_l|h) \log \frac{q(m_l|h)}{z(m_l)} dm_l\ dh \\
        &= \sum_l^L \mathds{E}_{h \sim p(h)} \left[ D_{KL} \left( q(m_l|h) || z(m_l)) \right) \right]
    \end{aligned}
\end{equation*}
where $z(m_l)$ is a standard Gaussian.
\end{proof}

\textbf{Proposition 5.1.} \textit{Utility mutual information is lower bounded by the contrastive NCE-binary objective, $I(M,Y) \geq \log \sigma (f(s,m,s_f^+)) +\log \sigma (1-f(s,m,s_f^-))$.}

\begin{proof}
We suppress the reliance on $h$ since this is directly passed through.
By definition of mutual information, we have,
\begin{equation*}
    \begin{aligned}
        I(M^j; Y^i) = \int \int p(m) \pi_{R^+}(y|m) \log \frac{\pi_{R^+}(y|m)}{\pi_{R^-}(y)} dm\,dy
    \end{aligned}
\end{equation*}
Our network model learns $\pi_{R^+}(y|m)$ from rolled-out trajectories, $R^+$, using our policy.
The prior of our network state, $\pi_{R^-}(y)$, can be modeled from rolling out a random trajectory, $R-$. 
Unfortunately, it is intractable to model $\pi_{R^+}(y|m)$ and $\pi_{R^-}(y)$ directly during iterative learning, but we can sample $y^+ \sim \pi_{R^+}(y|m)$ and $y^- \sim \pi_{R^-}(y)$ directly from our network during training.

It has been shown that $\log p(y|m)$ provides a bound on mutual information~\cite{poole2019variational},
\begin{equation}\label{eq:contrastive_MI}
    \begin{aligned}
        I(M^j;Y^i) \geq \mathds{E} \left[ \frac{1}{K} \sum_{k=1}^K \log \pi_{R^+}(y_k|m_k) + \log \pi_{R^-}(y_k) \right]
    \end{aligned}
\end{equation}
with the expectation over $\prod_l p(m_l, y_l)$. However, we need a tractable understanding of the information $Y$. 

\begin{lemma}\label{lemma:y}
$\pi_{R^-}(y) = p(s^\prime = s_f^- | y)$.
\end{lemma}
In the information bottleneck, $Y$ represents the desired outcome. In our setup, $y$ is coordination information that helps create the desired output, such as any action $a^-$. This implies, $y \implies a^-$. Since the transition is known, it follows that $a^- \implies s_f^-$
, a random future state. Thus, we have, $\pi_{R^-}(y) = p(s^\prime = s_f^- | y)$.

\begin{lemma}\label{lemma:ym}
$\pi_{R^+}(y|m) = p(s^\prime = s_f^+ | y,m)$.
\end{lemma}
This is similar to the proof for lemma~\ref{lemma:y}, but requires assumptions on messages $m$ from the emergent language. We note that when $m$ is random, the case defaults to lemma~\ref{lemma:y}. Thus, we assume we have at least input-oriented information in $m$ given sufficiently satisfying equation~\ref{eq:input}. Given a sufficient emergent language, it follows that $y \implies a^+$, where $a^+$ is an intention action based on $m$. Similarly, since the transition is known, $a^+ \implies s_f^+$, a desired goal state along the trajectory. Thus, we have, $\pi_{R^+}(y|m) = p(s^\prime = s_f^+ | y,m)$.

Recall the following (as shown in~\cite{eysenbach2022contrastive}), which we have adapted to our communication objective,
\begin{proposition}[rewards $\rightarrow$ probabilities]\label{eq:eysenbach}
The Q-function for the goal-conditioned reward function $r_g (s_t, m_t) = (1-\gamma) p(s^\prime = s_g |y_t)$ is equivalent to the probability of state $s_g$ under the discounted state occupancy measure:
\begin{equation}\label{eq:qfunc}
    Q_{s_g}^\pi (s,m) = p^{\pi} (s_f^+ = s_g |y)
\end{equation}
\end{proposition}
and
\begin{lemma}\label{lemma:eysenbach}
The critic function that optimizes equation~\ref{eq:contrastive_MI} is a Q-function for the goal-conditioned reward function up to a multiplicative constant $\frac{1}{p(s_f)}$: $\exp(f^*(s,m,s_f) = \frac{1}{p(s_f)} Q_{s_f}^\pi (s,m)$.
\end{lemma}
The critic function $f(s,m,s_f) = y^\intercal \texttt{enc}(s_f)$ represents the similarity between the encoding $y = \texttt{enc}(s,m)$ and the encoding of the future rollout $s_f$.

Given lemmas~\ref{lemma:y}~\ref{lemma:ym}~\ref{lemma:eysenbach} and proposition~\ref{eq:eysenbach}, it follows that equation~\ref{eq:contrastive_MI} is the NCE-binary~\cite{ma2018noise} (InfoMAX~\cite{hjelm2018learning}) objective,
\begin{equation}\label{eq:contrastive}
    \hat{I}(M^j,Y^i) = \log\left( \sigma(f(s,m,s_f^+)) \right) + \log\left(1 - \sigma(f(s,m,s_f^-)) \right)
\end{equation}
which lower bounds the mutual information, $I(M^j,Y^i) \geq \hat{I}(M^j,Y^i)$.
The critic function is unbounded, so we constrain it to $[0,1]$ with the sigmoid function, $\sigma(*)$.
We suppress the reliance on $h$ since this is directly passed through.
By definition of mutual information, we have,
\begin{equation*}
    \begin{aligned}
        I(M^j; Y^i) = \int \int p(m) \pi_{R^+}(y|m) \log \frac{\pi_{R^+}(y|m)}{\pi_{R^-}(y)} dm\,dy
    \end{aligned}
\end{equation*}
Our network model learns $\pi_{R^+}(y|m)$ from rolled-out trajectories, $R^+$, using our policy.
The prior of our network state, $\pi_{R^-}(y)$, can be modeled from rolling out a random trajectory, $R-$. 
Unfortunately, it is intractable to model $\pi_{R^+}(y|m)$ and $\pi_{R^-}(y)$ directly during iterative learning, but we can sample $y^+ \sim \pi_{R^+}(y|m)$ and $y^- \sim \pi_{R^-}(y)$ directly from our network during training.

It has been shown that $\log p(y|m)$ provides a bound on mutual information~\cite{poole2019variational},
\begin{equation}\label{eq:contrastive_MI}
    \begin{aligned}
        I(M^j;Y^i) \geq \mathds{E} \left[ \frac{1}{K} \sum_{k=1}^K \log \pi_{R^+}(y_k|m_k) + \log \pi_{R^-}(y_k) \right]
    \end{aligned}
\end{equation}
with the expectation over $\prod_l p(m_l, y_l)$. However, we need a tractable understanding of the information $Y$. 

\begin{lemma}\label{lemma:y}
$\pi_{R^-}(y) = p(s^\prime = s_f^- | y)$.
\end{lemma}
In the information bottleneck, $Y$ represents the desired outcome. In our setup, $y$ is coordination information that helps create the desired output, such as any action $a^-$. This implies, $y \implies a^-$. Since the transition is known, it follows that $a^- \implies s_f^-$
, a random future state. Thus, we have, $\pi_{R^-}(y) = p(s^\prime = s_f^- | y)$.

\begin{lemma}\label{lemma:ym}
$\pi_{R^+}(y|m) = p(s^\prime = s_f^+ | y,m)$.
\end{lemma}
This is similar to the proof for lemma~\ref{lemma:y}, but requires assumptions on messages $m$ from the emergent language. We note that when $m$ is random, the case defaults to lemma~\ref{lemma:y}. Thus, we assume we have at least input-oriented information in $m$ given sufficiently satisfying equation~\ref{eq:input}. Given a sufficient emergent language, it follows that $y \implies a^+$, where $a^+$ is an intention action based on $m$. Similarly, since the transition is known, $a^+ \implies s_f^+$, a desired goal state along the trajectory. Thus, we have, $\pi_{R^+}(y|m) = p(s^\prime = s_f^+ | y,m)$.

Recall the following (as shown in~\cite{eysenbach2022contrastive}), which we have adapted to our communication objective,
\begin{proposition}[rewards $\rightarrow$ probabilities]\label{eq:eysenbach}
The Q-function for the goal-conditioned reward function $r_g (s_t, m_t) = (1-\gamma) p(s^\prime = s_g |y_t)$ is equivalent to the probability of state $s_g$ under the discounted state occupancy measure:
\begin{equation}\label{eq:qfunc}
    Q_{s_g}^\pi (s,m) = p^{\pi} (s_f^+ = s_g |y)
\end{equation}
\end{proposition}
and
\begin{lemma}\label{lemma:eysenbach}
The critic function that optimizes equation~\ref{eq:contrastive_MI} is a Q-function for the goal-conditioned reward function up to a multiplicative constant $\frac{1}{p(s_f)}$: $\exp(f^*(s,m,s_f) = \frac{1}{p(s_f)} Q_{s_f}^\pi (s,m)$.
\end{lemma}
The critic function $f(s,m,s_f) = y^\intercal \texttt{enc}(s_f)$ represents the similarity between the encoding $y = \texttt{enc}(s,m)$ and the encoding of the future rollout $s_f$.

Given lemmas~\ref{lemma:y}~\ref{lemma:ym}~\ref{lemma:eysenbach} and proposition~\ref{eq:eysenbach}, it follows that equation~\ref{eq:contrastive_MI} is the NCE-binary~\cite{ma2018noise} (InfoMAX~\cite{hjelm2018learning}) objective,
\begin{equation}\label{eq:contrastive}
    \hat{I}(M^j,Y^i) = \log\left( \sigma(f(s,m,s_f^+)) \right) + \log\left(1 - \sigma(f(s,m,s_f^-)) \right)
\end{equation}
which lower bounds the mutual information, $I(M^j,Y^i) \geq \hat{I}(M^j,Y^i)$.
The critic function is unbounded, so we constrain it to $[0,1]$ with the sigmoid function, $\sigma(*)$.
\end{proof}






\end{document}